\documentclass[11pt]{article}

\usepackage[preprint]{acl}
\usepackage{algorithm}
\usepackage{algorithmic}
\usepackage[most]{tcolorbox}
\usepackage{authblk}
\usepackage{times}
\usepackage{latexsym}
\usepackage{subcaption}
\usepackage{colortbl}
\usepackage{xcolor}
\usepackage{multirow}
\usepackage{booktabs}
\usepackage{array}
\usepackage{enumitem}

\usepackage{amsmath}
\usepackage{amsfonts}
\usepackage[T1]{fontenc}

\usepackage[utf8]{inputenc}

\usepackage{microtype}

\usepackage{inconsolata}

\usepackage{graphicx}

\title{HMPO: Hybrid Median-length Policy Optimization for Chain-of-Thought Compression}

\author{
  \textbf{Minghui Zheng}\thanks{Equal contribution.},\ 
  \textbf{Hongxu Chen}$^{*}$,\ 
  \textbf{Huimin Ren}\thanks{Corresponding author.},\
  \textbf{Hongsheng Xin},\ 
  \textbf{Xiaoyang Qu} \\
  \textbf{Ze Wang},\ 
  \textbf{Shuling Yang},\ 
  \textbf{Ziyu Peng},\ 
  \textbf{Kaike Zhang},\ 
  \textbf{Zhou Pan},\ 
  \textbf{Kun Zhan} \\
  Li Auto Inc. \\
  \texttt{\{zhengminghui1, chenhongxu1, renhuimin\}@lixiang.com} \\
}

\begin{document}
\maketitle
\begin{abstract}
Large language models achieve remarkable performance via extended chain-of-thought (CoT) reasoning, yet this lengthy process incurs substantial inference overhead. Existing CoT compression methods struggle with inflexible manual length budgets, computationally expensive multi-stage training pipelines, and fragile scalability restricted to small models. We propose \textbf{HMPO} (\textbf{H}ybrid \textbf{M}edian-length \textbf{P}olicy \textbf{O}ptimization), a cost-effective, single-stage reinforcement learning framework. HMPO efficiently compresses CoT via three synergistic components: an \emph{adaptive median-based budget} derived from successful rollouts to eliminate manual tuning, a \emph{cosine-decay token reward} for smooth length penalization, and a \emph{multiplicative reward formulation} that substantially mitigates trivial reward hacking by strictly prioritizing answer correctness. Trained exclusively on mathematical data, HMPO generalizes seamlessly across math, code, science, and instruction-following tasks. Extensive experiments scaling from 9B to 122B parameters across dense and Mixture-of-Experts (MoE) architectures demonstrate that HMPO achieves \textbf{19\%--46\% token compression with negligible accuracy degradation}, all while drastically reducing training costs compared to existing multi-stage baselines.
\end{abstract}

\section{Introduction}
\begin{figure}[t]
    \centering
    \includegraphics[width=\linewidth]{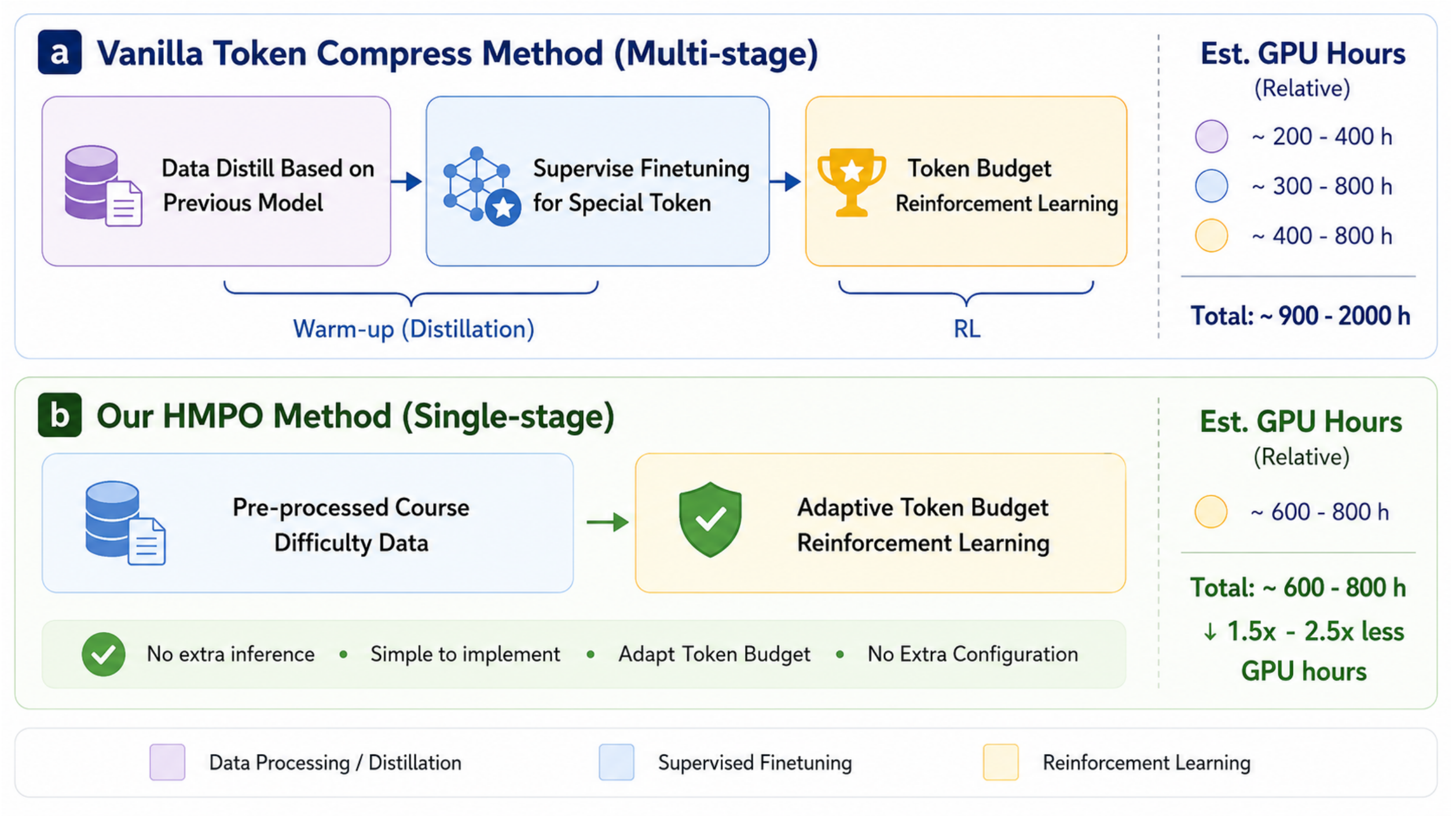}
    \caption{Comparison of training pipelines. Unlike vanilla multi-stage methods (a) that require data distillation and SFT warm-ups, our HMPO (b) achieves compression in a single RL stage, reducing GPU hours by 1.5$\times$--2.5$\times$.}
    \label{fig:intro_comparison}
\end{figure}

Recent large language models (LLMs) like OpenAI o1~\cite{jaech2024openai}, DeepSeek-R1~\cite{guo2025deepseek}, and Qwen-3~\cite{yang2025qwen3} achieve remarkable reasoning performance via extended chains of thought (CoT). While powerful across math, science, and coding tasks, this ``thinking'' paradigm incurs prohibitive costs: models routinely generate thousands of tokens per query, leading to high latency and resource consumption~\cite{chen2025not,sui2025stop}. This raises a pressing question: \emph{can we compress the reasoning process without sacrificing accuracy?}

Despite recent progress, existing CoT compression approaches share three critical limitations:
\textit{(1) Inflexible constraints and domain specificity:} Methods rely on manually tuned length budgets~\cite{wu2026art, lyu2025hierarchical} or rigid binary ``think or skip'' decisions~\cite{zhang-etal-2025-adaptthink}, limiting evaluations mostly to narrow mathematical domains.
\textit{(2) Prohibitive training costs:} Achieving effective compression typically requires computationally expensive pipelines, such as progressive multi-stage pruning~\cite{hou2025thinkprune}, data distillation with Supervised Fine-Tuning (SFT) warm-ups~\cite{fang2026thinkless}, or massive pre-sampling~\cite{luo2025o1}.
\textit{(3) Fragile scalability:} Current evaluations are restricted to small dense models ($\leq$ 32B)~\cite{fan2026ctrlcot}, leaving applicability to state-of-the-art large-scale and Mixture-of-Experts (MoE) architectures unverified.

To address these limitations, we propose \textbf{HMPO} (\textbf{H}ybrid \textbf{M}edian-length \textbf{P}olicy \textbf{O}ptimization), a cost-effective, single-stage reinforcement learning framework (\textbf{Figure~\ref{fig:intro_comparison}}). Built upon  Group Relative Policy Optimization (GRPO)~\cite{shao2024deepseekmath}, HMPO elegantly compresses CoT through three synergistic components: 
(1) an \emph{adaptive median-based budget} that dynamically targets the median length of \emph{correct} rollouts, implicitly forming a self-tightening curriculum without manual tuning; 
(2) a \emph{cosine-decay token reward} applied exclusively to successful trajectories, providing a ``soft landing'' penalty that compresses lengths while preserving necessary reasoning steps; and 
(3) a \emph{multiplicative reward formulation} ($R = R_{\text{acc}} \cdot R_{\text{token}}$) that enforces a strict ``correctness-first, length-second'' hierarchy, mathematically inoculating the training against reward hacking by denying efficiency gradients to incorrect responses.

As illustrated in Figure~\ref{fig:intro_comparison}, by eliminating distillation and SFT phases, HMPO reduces training GPU hours by 1.5$\times$--2.5$\times$ over multi-stage baselines. Crucially, it unlocks massive inference token savings. Trained solely on math data, HMPO generalizes seamlessly across math, code, science, and instruction-following tasks. Scaling from 9B to 122B parameters across dense and MoE architectures, it achieves \textbf{19\%--46\% token compression with negligible accuracy degradation}. Our main contributions are:
\begin{itemize}
    \item \textbf{Methodological Innovation:} HMPO introduces adaptive correct-median budgeting, cosine reward shaping, and multiplicative reward factorization, jointly resolving the budget rigidity, penalty aggressiveness, and reward hacking vulnerabilities of prior RL-based compression methods.
    \item \textbf{Cross-domain Generalization:} We present the most extensive CoT compression evaluation to date, validating HMPO across 9B to 122B dense and MoE models, demonstrating robust cross-domain generalization despite math-only training.
    \item \textbf{Unprecedented Scalability to 100B+ Models:} We present the first successful CoT compression on massive MoE architectures. On the state-of-the-art 122B model, HMPO achieves highly compact reasoning while maintaining near-perfect accuracy parity. This enables our models to perform competitively against substantially larger models---matching or outperforming frontier models of much larger activated scales (e.g., 700B+ MoEs) with drastically fewer tokens.
\end{itemize}

\section{Related Work}
\label{sec:related}
\paragraph{Efficient Chain-of-Thought Reasoning.}
While Chain-of-Thought (CoT)~\cite{wei2022chain} equips reasoning models like OpenAI o1~\cite{jaech2024openai} and DeepSeek-R1~\cite{guo2025deepseek} with advanced capabilities, it often induces severe inference latency via ``overthinking''~\cite{chen2025not,sui2025stop}. To mitigate this, recent efficient reasoning approaches explore supervised fine-tuning on compressed data~\cite{munkhbat2025self, kang2025c3ot, xia2025tokenskip}, post-hoc pruning~\cite{zeng2025pruning}, and length-aware reinforcement learning~\cite{aggarwal2025l1, yeo2025demystifying}.

\paragraph{CoT Compression via RL and Adaptive Reasoning.}
The most closely related works fall into length-aware RL, iterative pruning, and adaptive reasoning. Length-aware RL methods~\cite{aggarwal2025l1, lyu2025hierarchical, han2025token, arora2026training} incorporate explicit length penalties but strictly rely on manually predefined or statically estimated token budgets.  Iterative pruning methods~\cite{hou2025thinkprune, luo2025o1} and adaptive ``think-or-skip'' frameworks such as AdaptThink~\cite{zhang-etal-2025-adaptthink} and Thinkless~\cite{fang2026thinkless} conditionally reduce CoT length but necessitate computationally prohibitive multi-stage pipelines or data distillation. In contrast, HMPO seamlessly integrates fine-grained length shaping into a \emph{single RL stage}, entirely replacing rigid manual constraints with an \emph{on-policy adaptive median budget}.

\section{Method}
\label{sec:method}

\begin{figure*}[ht]
    \centering
    \includegraphics[width=\linewidth]{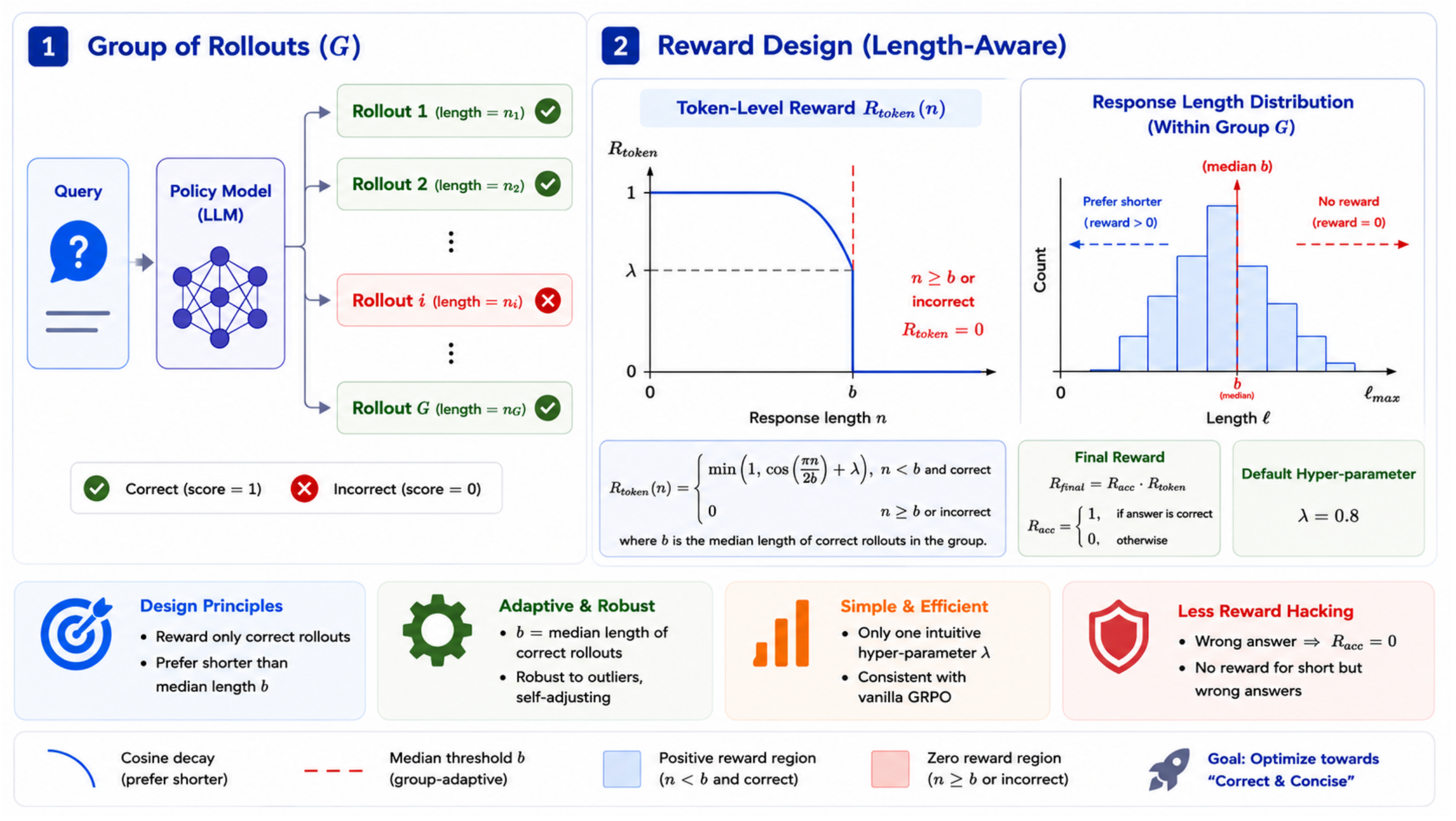}
    \caption{Overview of HMPO. \textbf{Left:} For each query, the policy samples a group of rollouts ($G$). \textbf{Right:} Instead of relying on a static threshold, HMPO dynamically derives an adaptive budget $b$ from the median length of only the \emph{correct} rollouts to construct a smooth cosine-decay token reward. \textbf{Bottom:} The final reward is combined multiplicatively to enforce a strict ``correctness-first, length-second'' objective, mathematically preventing reward hacking (i.e., short but incorrect answers strictly receive zero reward).}
    \label{fig:method}
\end{figure*}

As illustrated in Figure~\ref{fig:method}, our proposed HMPO builds upon GRPO~\cite{shao2024deepseekmath} by dynamically deriving a group-adaptive budget $b$ from the median length of correct rollouts and applying a length-aware token reward. 

\subsection{Preliminaries}
\label{sec:method_prelim}
For each query $(q,a) \sim \mathcal{D}$, the policy $\pi_\theta$ generates a response $o = (o_1, \dots, o_{|o|})$ of $|o|$ tokens. Following GRPO, we sample $G$ rollouts $\{o_i\}_{i=1}^G$ from the old policy $\pi_{\theta_{\mathrm{old}}}(\cdot \mid q)$ to optimize the clipped surrogate objective:
\begin{equation}
\mathcal{J}_{\mathrm{GRPO}}(\theta)
=
\mathbb{E}_{(q,a) \sim \mathcal{D}}\!\left[
\frac{1}{G}\sum_{i=1}^G
\frac{1}{|o_i|}
\sum_{t=1}^{|o_i|}
\ell_{i,t}(\theta)
\right]
\label{eq:grpo}
\end{equation}
where the token-level importance ratio $s_{i,t}(\theta)$ is defined as:
\begin{equation}
s_{i,t}(\theta)
=
\frac{\pi_\theta(o_{i,t}\mid q,o_{i,<t})}
{\pi_{\theta_{\mathrm{old}}}(o_{i,t}\mid q,o_{i,<t})}
\label{eq:ratio}
\end{equation}
and the token-level surrogate loss $\ell_{i,t}(\theta)$ is split across lines to fit the column width:
\begin{equation}
\begin{aligned}
\ell_{i,t}(\theta) = \min \Big( & s_{i,t}(\theta) A_i, \\
& \text{clip}\big(s_{i,t}(\theta), 1-\epsilon, 1+\epsilon\big) A_i \Big)
\end{aligned}
\label{eq:grpo_loss}
\end{equation}
The group-relative advantage $A_i$ for each rollout $o_i$ is computed as:
\begin{equation}
A_i
=
\frac{
R_i - \text{mean}(\{R_j\}_{j=1}^G)
}{
\text{std}(\{R_j\}_{j=1}^G) + \varepsilon}
\label{eq:advantage}
\end{equation}
where $R_i$ denotes the rollout-level reward for sample $o_i$, and $\varepsilon$ is a small constant added for numerical stability.

Standard GRPO rewards focus on answer correctness. To mitigate the high inference costs of lengthy CoT, we must additionally optimize for reasoning efficiency. However, penalizing verbosity without degrading accuracy requires establishing an appropriate token budget. Therefore, our core modification to GRPO lies in redesigning the rollout reward $R_i$ to jointly enforce this budget-aware efficiency and answer correctness. where $R_i$ is the rollout-level reward for sample $o_i$. 
\subsection{Adaptive Budget Estimation}
\label{sec:method_budget}
In RL-based CoT compression, efficiency is typically achieved by incorporating length-penalizing rewards, which fundamentally rely on defining a target token budget~\cite{lyu2025hierarchical, hou2025thinkprune,wu2026art}. However, establishing an appropriate budget remains a critical challenge. Existing paradigms typically resort to either a dataset-wide static budget, which fails to accommodate varying problem complexities, or prior rollout-based estimation, which incurs prohibitive computational overhead and tightly couples the algorithm to pre-computed data.

To overcome these bottlenecks, HMPO introduces an \textbf{adaptive median-length budget} calculated dynamically from on-policy valid reasoning traces. Let $\{o_i\}_{i=1}^G$ be the current rollout group with respective token lengths $n_i = |o_i|$, and let $\mathcal{C} \subseteq \{1, \dots, G\}$ denote the indices of rollouts that yield the correct answer. The group-level budget $b$ is formally defined as the median length of these successful attempts:
\begin{equation}
b = \mathrm{median}(\{n_i \mid i \in \mathcal{C}\})
\label{eq:budget}
\end{equation}

This formulation is mathematically simple yet highly effective. Statistically, computing the median exclusively on correct rollouts isolates the length signal from noisy, failed reasoning attempts and provides strong robustness against the heavy-tailed distribution of LLM generation. Dynamically, it perfectly decouples the budget from data preprocessing while enabling a difficulty-aware constraint: complex queries intrinsically produce longer correct rollouts, naturally relaxing the budget, whereas simpler queries are subjected to tighter constraints. Furthermore, as the policy learns to reason more efficiently over time, $b$ automatically decreases, implicitly forming a self-tightening curriculum without manual intervention.

\subsection{Token-aware Reward Design}
\label{sec:method_reward}

With the dynamic budget $b$ in place, the remaining question is how to incorporate it into the optimization objective. A naive length penalty is prone to reward hacking, where the model produces overly
short but incorrect responses to exploit the length signal. To prevent this, we factorize the overall reward into an accuracy term and a token-efficiency term, enforcing a strict \emph{correctness-first, length-second} hierarchy.

\paragraph{Accuracy reward.} $R_{\mathrm{acc}}$ measures whether the final prediction is correct. In our experiments, it is assigned as $1$ for correct and $0$ for incorrect.

\paragraph{Token reward.} To steer the policy toward ``correct-and-concise'' behavior with minimal deviation from standard on-policy updates, we introduce a length preference \emph{exclusively} for successful trajectories. Specifically, we define $R_{\mathrm{token}}$ to reinforce correct rollouts whose length $n$ is below the adaptive budget $b$, applying a smooth cosine decay:

{\footnotesize
\begin{equation}
R_{\mathrm{token}} =
\begin{cases}
\min\!\big(
1,\;
 \cos\!\left(\tfrac{\pi n}{2b}\right) + \lambda
\big),
& \text{if correct and } n < b, \\[6pt]
0, & \text{otherwise},
\end{cases}
\label{eq:token_reward}
\end{equation}
}
where the reward is applied only to correct trajectories with $n > 0$; all incorrect or empty ($n=0$) responses receive $R_{\mathrm{token}} = 0$. The scalar $\lambda$ controls the magnitude of the length-induced gain. The cosine term guarantees a smooth, bounded preference: shorter correct responses receive higher rewards, with the signal decaying gently as $n$ approaches $b$. The $\min(1,\cdot)$ truncation ensures the reward does not over-dominate the accuracy signal.

\paragraph{Final reward.} We combine the accuracy and token efficiency signals multiplicatively:
\begin{equation}
R_{\mathrm{final}} = R_{\mathrm{acc}} \cdot R_{\mathrm{token}}
\label{eq:final_reward}
\end{equation}
Consider four trajectory categories: \textit{correct-short}, \textit{correct-long}, \textit{incorrect-short}, \textit{incorrect-long}. Under this formulation, only correct-and-concise trajectories receive high reward; incorrect and correct-but-beyond-budget ones are zeroed out ($R_{\mathrm{final}}=0$). In contrast, additive combination ($R_{\mathrm{acc}} + R_{\mathrm{token}} \in [0,2]$) still grants correct-but-long trajectories a base reward of $R_{\mathrm{acc}}=1$ regardless of verbosity, making the optimization signal less direct. The multiplicative form provides a sharper target and reduces reward misalignment---this motivates our default choice.

\subsection{Training Procedure}
\label{sec:method_training}
Algorithm~\ref{alg:hmpo} summarizes the overall HMPO training loop. For each input query, we sample a group of responses from the old policy. Crucially, we first compute the accuracy reward for each rollout to identify the correct subset $\mathcal{C}$. We then estimate the adaptive budget $b$ using the median length of these valid traces. Finally, we calculate $R_{\mathrm{token}}$ and the multiplicative $R_{\mathrm{final}}$ to construct normalized advantages, which are subsequently plugged into the GRPO objective for policy optimization.

\begin{algorithm}[tb] 
\small
\caption{HMPO Training}
\label{alg:hmpo}
\begin{algorithmic}[1]
\REQUIRE Dataset $\mathcal{D}$, policy $\pi_\theta$, old policy $\pi_{\theta_{\mathrm{old}}}$, group size $G$, penalty coefficient $\lambda$
\WHILE{not converged}
    \STATE Sample a batch of queries $\{q\} \sim \mathcal{D}$
    \FOR{each query $q$}
        \STATE Sample $G$ rollouts $\{o_i\}_{i=1}^G \sim \pi_{\theta_{\mathrm{old}}}(\cdot \mid q)$
        \FOR{$i = 1$ to $G$}
            \STATE Compute length $n_i = |o_i|$ and accuracy reward $R_{\mathrm{acc}}^{(i)}$
        \ENDFOR
        \STATE Identify correct subset $\mathcal{C} = \{i \mid R_{\mathrm{acc}}^{(i)} > 0\}$
        \IF{$|\mathcal{C}| > 0$}
            \STATE Estimate adaptive budget $b \leftarrow \mathrm{median}(\{n_i \mid i \in \mathcal{C}\})$
        \ELSE
            \STATE $b \leftarrow \infty$ \COMMENT{Fallback; $R_{\mathrm{final}}$ will be $0$ regardless}
        \ENDIF
        \FOR{$i = 1$ to $G$}
            \STATE Compute token reward $R_{\mathrm{token}}^{(i)}$ using Eq.~\ref{eq:token_reward}
            \STATE Compute final reward $R_i \leftarrow R_{\mathrm{acc}}^{(i)} \cdot R_{\mathrm{token}}^{(i)}$
        \ENDFOR
        \STATE Normalize $\{R_i\}_{i=1}^G$ into advantages $\{A_i\}_{i=1}^G$
    \ENDFOR
    \STATE Update $\pi_\theta$ using the GRPO objective in Eq.~\ref{eq:grpo}
\ENDWHILE
\end{algorithmic}
\end{algorithm}







\section{Experiments}

\label{sec:experiments}

\subsection{Experimental Setup}
\label{sec:setup}

\paragraph{Models, Datasets, and Baselines.}
To evaluate scalability across architectures and model sizes, we utilize Qwen3.5-9B~\cite{qwen3.5} as our dense model, alongside Qwen3.5-35B-A3B~\cite{qwen3.5} and Qwen3.5-122B-A10B~\cite{qwen3.5} as Mixture-of-Experts (MoE) models. 
We train on 6.5K challenging instances (levels 5--9) sampled from DeepMath-103K~\cite{he2025deepmath103k}. 
We compare HMPO against the original Base Models and three CoT compression baselines: \textbf{AdaptThink}~\cite{zhang-etal-2025-adaptthink} (constrained binary mode selection), \textbf{Thinkless}~\cite{fang2026thinkless} (two-stage SFT and decoupled RL), and \textbf{ThinkPrune}~\cite{hou2025thinkprune} (multi-round iterative length pruning). To ensure fair comparisons, all baselines are rigorously reproduced using their original training pipelines and hyperparameters.Detailed reproduction procedures for all baselines are provided in Appendix~\ref{app:baseline_reproduction}. All artifacts (models, datasets, and codebases) utilized in this work are publicly available and employed in strict compliance with their open-source licenses (e.g., Apache 2.0, MIT). We will release our code, training scripts, and configurations upon acceptance.

\paragraph{Evaluation Metrics.}
We evaluate cross-domain performance on five benchmarks: AIME 2025~\cite{aime25} and AIME 2026~\cite{aime26} (math), GPQA-Diamond~\cite{rein2023gpqa} (science), IFEval~\cite{zhou2023instructionfollowingevaluationlargelanguage} (instruction following), and LiveCodeBench-V6~\cite{jain2025livecodebench} (code). 
For each benchmark, we report accuracy (Acc), average token length (Len), and per-benchmark compression ratio ($\rho$). We also report the overall cross-benchmark compression ratio $\bar{\rho}$, defined as:
\begin{equation}
    \bar{\rho} = \frac{\sum_{i=1}^{5} \text{Len}_{\text{base}}^{(i)} - \sum_{i=1}^{5} \text{Len}_{\text{method}}^{(i)}}{\sum_{i=1}^{5} \text{Len}_{\text{base}}^{(i)}}.
    \label{eq:overall_compression}
\end{equation}
To maximally preserve the models' inherent capabilities and ensure robust performance under token compression, inference is conducted via SGLang~\cite{zheng2024sglang} using official recommended settings (temperature $1.0$, top-$p{=}0.95$, top-$k{=}20$) and a full 256K maximum context limit. To reduce variance, AIME results are averaged over 16 sampled responses per problem, and other benchmarks are evaluated twice.

\paragraph{Implementation Details.}
We implement HMPO using the VERL framework~\cite{sheng2025hybridflow}, sampling $G{=}10$ rollouts per prompt with $\lambda{=}0.8$. Training converges within 40--60 steps, reaching the optimal accuracy-compression trade-off around steps 16--24. Full hyperparameters and infrastructure details are deferred to Appendix~\ref{app:hyperparams}.

\subsection{Main Results}

\begin{table*}[!t]
\centering
\scriptsize
\setlength{\tabcolsep}{3pt}
\renewcommand{\arraystretch}{0.95}
\caption{Main results on three Qwen3.5-series model configurations. Each method occupies two rows: accuracy (Acc, \%) and output token length (Len). Green/red annotations indicate improvement/degradation relative to the Base Model. The Avg column shows average accuracy and average length (with overall compression ratio $\bar{\rho}$). \textbf{Bold}: HMPO results in the Avg column. $^{\dag}$: baselines not reproduced due to training cost.
}
\label{tab:main_results}
\resizebox{\textwidth}{!}{
\begin{tabular}{ll l l l l l l}
\specialrule{2pt}{0pt}{0pt}
\noalign{\vskip 2pt}
\textbf{Method} & & \textbf{AIME'25} & \textbf{AIME'26} & \textbf{LCB-V6} & \textbf{GPQA} & \textbf{IFEval} & \textbf{Avg.} \\
\midrule
\rowcolor{gray!20} \multicolumn{8}{c}{\textbf{Qwen3.5-9B} \rule{0pt}{6pt}} \\
\multirow{2}{*}{Base Model}
  & Acc & 85.21 & 89.90 & 65.60 & 81.57 & 91.82 & 82.82 \\
  & Len & 24,947 & 24,871 & 49,227 & 10,586 & 5,692 & 23,065 \\
\addlinespace[1.5pt]
\multirow{2}{*}{AdaptThink}
  & Acc & 81.25 & 87.29 & 66.79 & 82.58 & 92.42 & 82.07 \textcolor{red!50!black}{\scriptsize -0.75} \\
  & Len & 19,797 \textcolor{green!50!black}{\scriptsize -21\%} & 20,548 \textcolor{green!50!black}{\scriptsize -17\%} & 35,577 \textcolor{green!50!black}{\scriptsize -28\%} & 8,366 \textcolor{green!50!black}{\scriptsize -21\%} & 4,468 \textcolor{green!50!black}{\scriptsize -22\%} & 17,751 \textcolor{green!50!black}{\scriptsize -23\%} \\
\addlinespace[1.5pt]
\multirow{2}{*}{Thinkless}
  & Acc & 13.75 & 8.33 & 8.78 & 1.26 & 20.24 & 10.47 \textcolor{red!50!black}{\scriptsize -72.35} \\
  & Len & 560 \textcolor{green!50!black}{\scriptsize -98\%} & 604 \textcolor{green!50!black}{\scriptsize -98\%} & 604 \textcolor{green!50!black}{\scriptsize -99\%} & 548 \textcolor{green!50!black}{\scriptsize -95\%} & 1,239 \textcolor{green!50!black}{\scriptsize -78\%} & 711 \textcolor{green!50!black}{\scriptsize -97\%} \\
\addlinespace[1.5pt]
\multirow{2}{*}{ThinkPrune}
  & Acc & 73.75 & 77.71 & 56.11 & 80.81 & 89.65 & 75.61 \textcolor{red!50!black}{\scriptsize -7.21} \\
  & Len & 12,171 \textcolor{green!50!black}{\scriptsize -51\%} & 10,586 \textcolor{green!50!black}{\scriptsize -57\%} & 12,592 \textcolor{green!50!black}{\scriptsize -74\%} & 4,484 \textcolor{green!50!black}{\scriptsize -58\%} & 2,071 \textcolor{green!50!black}{\scriptsize -64\%} & 8,381 \textcolor{green!50!black}{\scriptsize -64\%} \\
\addlinespace[1.5pt]
\multirow{2}{*}{HMPO}
  & Acc & 84.58 & 89.79 & 65.27 & 82.32 & 92.05 & \textbf{82.80 \textcolor{red!50!black}{\scriptsize -0.02}} \\
  & Len & 17,087 \textcolor{green!50!black}{\scriptsize -32\%} & 15,564 \textcolor{green!50!black}{\scriptsize -37\%} & 19,283 \textcolor{green!50!black}{\scriptsize -61\%} & 6,934 \textcolor{green!50!black}{\scriptsize -34\%} & 3,366 \textcolor{green!50!black}{\scriptsize -41\%} & \textbf{12,447 \textcolor{green!50!black}{\scriptsize -46\%}} \\
\midrule
\rowcolor{gray!20} \multicolumn{8}{c}{\textbf{Qwen3.5-35B-A3B} \rule{0pt}{6pt}} \\
\multirow{2}{*}{Base Model}
  & Acc & 89.58 & 91.87 & 73.66 & 83.08 & 92.79 & 86.20 \\
  & Len & 18,900 & 19,920 & 31,024 & 7,107 & 7,365 & 16,863 \\
\addlinespace[1.5pt]
\multirow{2}{*}{AdaptThink}
  & Acc & 86.67 & 86.25 & 68.70 & 84.09 & 92.33 & 83.61 \textcolor{red!50!black}{\scriptsize -2.59} \\
  & Len & 12,897 \textcolor{green!50!black}{\scriptsize -32\%} & 13,470 \textcolor{green!50!black}{\scriptsize -32\%} & 25,851 \textcolor{green!50!black}{\scriptsize -17\%} & 6,708 \textcolor{green!50!black}{\scriptsize -6\%} & 9,761 \textcolor{red!50!black}{\scriptsize +33\%} & 13,737 \textcolor{green!50!black}{\scriptsize -19\%} \\
\addlinespace[1.5pt]
\multirow{2}{*}{Thinkless}
  & Acc & 60.42 & 67.92 & 52.29 & 77.27 & 69.13 & 65.41 \textcolor{red!50!black}{\scriptsize -20.79} \\
  & Len & 7,672 \textcolor{green!50!black}{\scriptsize -59\%} & 7,433 \textcolor{green!50!black}{\scriptsize -63\%} & 9,343 \textcolor{green!50!black}{\scriptsize -70\%} & 4,345 \textcolor{green!50!black}{\scriptsize -39\%} & 2,713 \textcolor{green!50!black}{\scriptsize -63\%} & 6,301 \textcolor{green!50!black}{\scriptsize -63\%} \\
\addlinespace[1.5pt]
\multirow{2}{*}{ThinkPrune}
  & Acc & 67.71 & 78.54 & 60.31 & 81.57 & 92.98 & 76.22 \textcolor{red!50!black}{\scriptsize -9.98} \\
  & Len & 5,798 \textcolor{green!50!black}{\scriptsize -69\%} & 5,295 \textcolor{green!50!black}{\scriptsize -73\%} & 7,815 \textcolor{green!50!black}{\scriptsize -75\%} & 2,806 \textcolor{green!50!black}{\scriptsize -61\%} & 1,664 \textcolor{green!50!black}{\scriptsize -77\%} & 4,676 \textcolor{green!50!black}{\scriptsize -72\%} \\
\addlinespace[1.5pt]
\multirow{2}{*}{HMPO}
  & Acc & 86.46 & 90.83 & 73.66 & 82.58 & 94.09 & \textbf{85.52 \textcolor{red!50!black}{\scriptsize -0.68}} \\
  & Len & 12,454 \textcolor{green!50!black}{\scriptsize -34\%} & 12,457 \textcolor{green!50!black}{\scriptsize -37\%} & 15,762 \textcolor{green!50!black}{\scriptsize -49\%} & 4,981 \textcolor{green!50!black}{\scriptsize -30\%} & 2,836 \textcolor{green!50!black}{\scriptsize -61\%} & \textbf{9,698 \textcolor{green!50!black}{\scriptsize -42\%}} \\
\midrule
\rowcolor{gray!20} \multicolumn{8}{c}{\textbf{Qwen3.5-122B-A10B}$^{\dag}$ \rule{0pt}{6pt}} \\
\multirow{2}{*}{Base Model}
  & Acc & 89.58 & 92.71 & 78.90 & 86.60 & 93.40 & 88.24 \\
  & Len & 17,666 & 17,072 & 23,774 & 6,871 & 4,016 & 13,880 \\
\addlinespace[1.5pt]
\multirow{2}{*}{HMPO}
  & Acc & 90.21 & 92.50 & 79.01 & 85.61 & 93.35 & \textbf{88.14 \textcolor{red!50!black}{\scriptsize -0.10}} \\
  & Len & 13,675 \textcolor{green!50!black}{\scriptsize -23\%} & 13,136 \textcolor{green!50!black}{\scriptsize -23\%} & 20,077 \textcolor{green!50!black}{\scriptsize -16\%} & 5,671 \textcolor{green!50!black}{\scriptsize -17\%} & 3,377 \textcolor{green!50!black}{\scriptsize -16\%} & \textbf{11,187 \textcolor{green!50!black}{\scriptsize -19\%}} \\
\specialrule{2pt}{0pt}{0pt}
\end{tabular}}
\end{table*}
\label{sec:main_results}

Table~\ref{tab:main_results} presents the overall comparison across three Qwen3.5-series models. Detailed metric layouts are described in the table caption. Notably, due to prohibitive computational costs, we omit baseline reproductions for the 122B configuration.

\paragraph{HMPO achieves the best accuracy-compression trade-off.}
HMPO consistently delivers substantial compression while preserving Base Model accuracy. On Qwen3.5-9B, HMPO nearly halves output tokens ($\bar{\rho}{=}46\%$) with a negligible 0.02 percentage points(pp) average accuracy drop. On Qwen3.5-35B-A3B, it compresses 42\% while maintaining 85.52\% average accuracy, even improving IFEval (94.09\% vs.\ 92.79\%). In contrast, ThinkPrune achieves higher compression (64--72\%) but suffers severe accuracy degradation (11--22pp AIME drops). Thinkless exhibits catastrophic failure on 9B ($<$15\% accuracy), likely due to SFT-induced mode collapse during the subsequent RL phase. AdaptThink provides marginal gains but surprisingly inflates 35B IFEval lengths by 33\%. Thus, HMPO uniquely occupies the Pareto frontier where both accuracy and efficiency are jointly optimized.

\paragraph{Consistent in-domain retention and out-of-domain generalization.}
Trained solely on math data, HMPO transfers its concise reasoning patterns seamlessly. On in-domain \textbf{AIME 2025/2026}, it secures 23--37\% compression, even boosting 122B accuracy on AIME 2025 (89.58\%$\to$90.21\%). Crucially, on \textit{out-of-domain} tasks, HMPO achieves 16--61\% compression with accuracy parity on \textbf{LiveCodeBench(LCB)-V6} (e.g., exactly matching the 35B Base Model's 73.66\% while cutting outputs by 49\%). It further yields 17--34\% compression on \textbf{GPQA-Diamond} without accuracy loss, and 16--61\% reduction on \textbf{IFEval} while maintaining or improving accuracy. This proves HMPO generalizes robustly without task-specific tuning.Compression case studies are presented in Appendix~\ref{app:aime_case_study}.

\paragraph{Robustness across model scales and architectures.}
HMPO remains effective scaling from 9B dense to 122B MoE models. On the 122B model, it yields 19\% compression with a mere 0.10pp average accuracy degradation. Crucially, HMPO requires \textit{no hyperparameter tuning}: the identical $\lambda{=}0.8$ setting works across all scales. By automatically calibrating compression pressure via the adaptive median budget, HMPO serves as a highly practical, drop-in optimization for deployment.


\begin{figure}[h]
    \centering
    \includegraphics[width=\columnwidth]{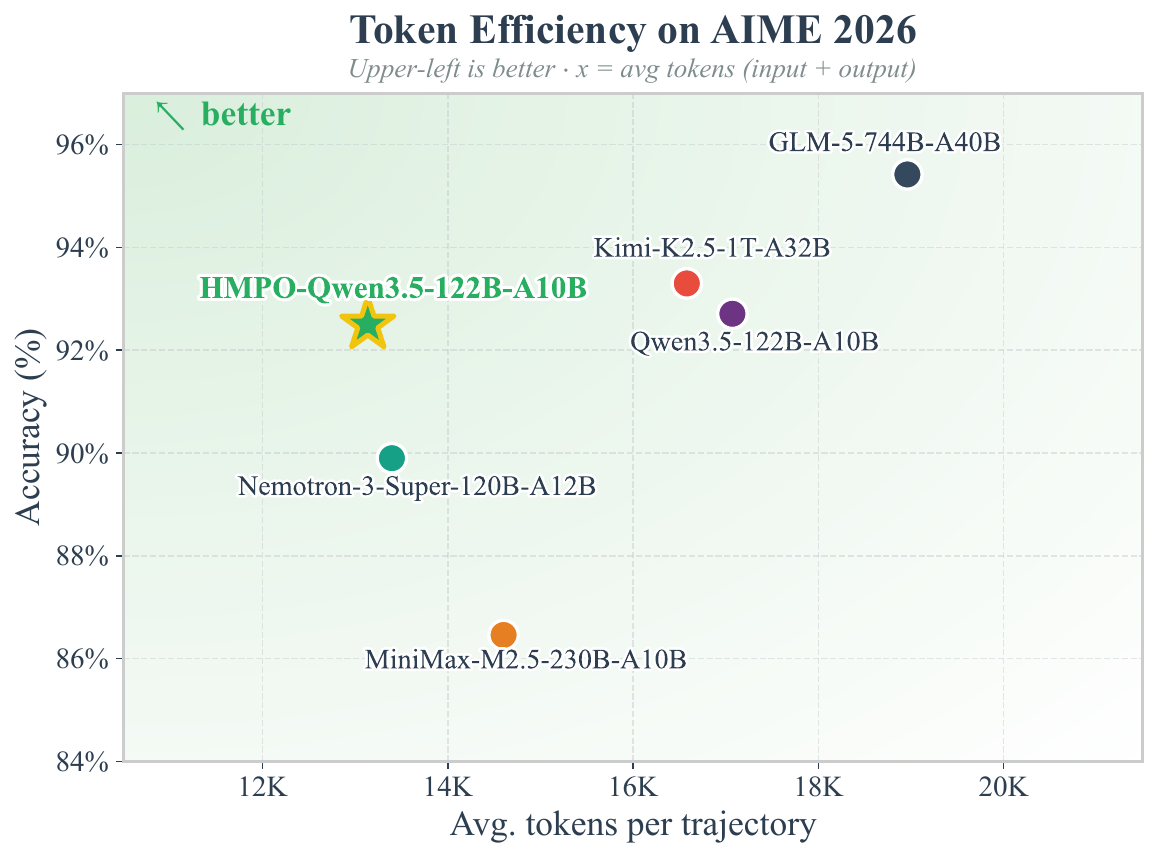} 
    \caption{Token efficiency on AIME 2026. Each point represents a model's accuracy vs.\ average tokens per trajectory (input + output). Upper-left is better. HMPO-trained models (stars) achieve competitive or superior accuracy with significantly fewer tokens than frontier models of much larger activated scale.}
    \label{fig:token_efficiency_aime}
\end{figure}
\paragraph{Token efficiency vs.\ frontier models.}
To contextualize HMPO's practical value, Figure~\ref{fig:token_efficiency_aime} plots accuracy versus average tokens per trajectory on AIME 2026. First, compared to its unoptimized base model, HMPO-Qwen3.5-122B-A10B slashes average token consumption by 23\% (17.1K $\to$ 13.1K) with a negligible 0.21\% accuracy variance (92.71\% vs.\ 92.50\%). Beyond self-improvement, HMPO fundamentally reshapes the Pareto frontier against substantially larger frontier models. It decisively outperforms Nemotron-3-Super-120B-A12B (89.90\% at 13.4K) and delivers performance highly competitive with massive models like Kimi-K2.5-1T-A32B (93.30\% at 16.6K). This demonstrates that HMPO enables reasoning models to punch significantly above their weight class, delivering top-tier mathematical performance at a fraction of the frontier inference cost.
\begin{figure*}[ht]
    \centering
    \includegraphics[width=\textwidth]{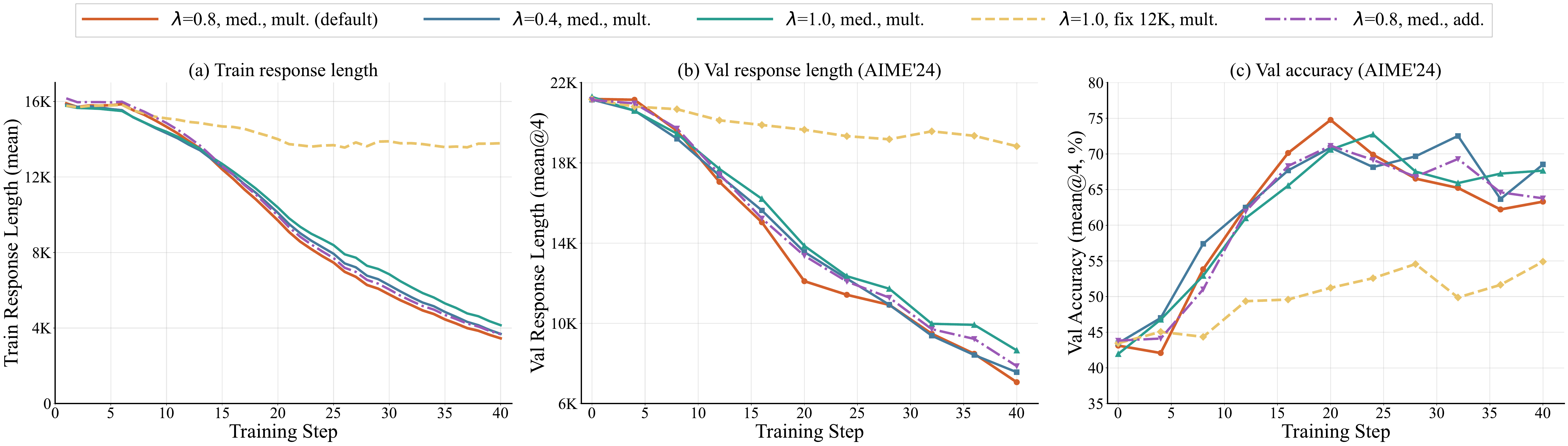}
    \caption{Ablation training dynamics on Qwen3.5-9B. Five configurations are compared: varying $\lambda$ (0.4, 0.8, 1.0), budget strategy (median vs.\ fixed 12K), and reward composition (multiplicative vs.\ additive). The default ($\lambda{=}0.8$, median, mult.) achieves the best accuracy-compression trade-off.}
    \label{fig:ablation_curves}
\end{figure*}

\subsection{Ablation Study}
\label{sec:ablation}
We ablate HMPO's components on Qwen3.5-9B (Table~\ref{tab:ablation}, Figure~\ref{fig:ablation_curves}). \textit{Note on Validation:} We monitor dynamics on AIME 2024. Since early-step outputs often exceed the 24K limit and truncate, initial accuracy is artificially deflated. As compression progresses, truncation ceases, revealing genuine accuracy before over-compression causes degradation. Thus, we emphasize relative convergence trends over absolute peaks.
\begin{table}[h]
\centering
\small
\setlength{\tabcolsep}{3pt}
\caption{Ablation study on Qwen3.5-9B, evaluated on all five benchmarks. We vary one component at a time from the default. All variants at step 16 except fixed-budget (step 40, due to slower convergence).}
\label{tab:ablation}

\begin{tabular}{@{}l@{\hspace{6pt}}r@{\hspace{6pt}}r@{}}
\toprule
\textbf{Variant} & \textbf{Avg Acc (\%)} & $\bar{\rho}$ \textbf{(\%)} \\
\midrule
\textbf{Default ($\lambda$=0.8, med., mult.)}  & 82.80  & 46.0 \\
\midrule
\multicolumn{3}{@{}l}{\textit{Compression strength}} \\
\quad $\lambda$=0.4, med., mult.    & 79.79  & 41.6 \\
\quad $\lambda$=1.0, med., mult.    & 81.15  & 42.5 \\
\midrule
\multicolumn{3}{@{}l}{\textit{Budget strategy}} \\
\quad $\lambda$=1.0, fixed 12K, mult. & 82.12  & 12.4 \\
\midrule
\multicolumn{3}{@{}l}{\textit{Reward composition}} \\
\quad $\lambda$=0.8, med., additive & 80.83  & 41.7 \\
\bottomrule
\end{tabular}
\end{table}

\paragraph{Adaptive median budget vs.\ fixed budget.}
We compare our adaptive median against a meticulously calibrated fixed budget ($b{=}12\text{K}$, set at 75\% of the Base Model's ${\sim}16\text{K}$ mean training-time response length, with $\lambda{=}1.0$). As shown in Figure~\ref{fig:ablation_curves}(a), the fixed budget fails to drive meaningful compression (stalling at ${\sim}14\text{K}$), as many responses already satisfy the threshold and receive no optimization signal. Conversely, the adaptive median dynamically self-calibrates to the batch's length distribution, forming a robust self-tightening curriculum that continuously compresses outputs ($16\text{K} \to 4\text{K}$). This avoids the sluggish convergence or premature collapse inherent to static manual tuning.

\paragraph{Multiplicative vs.\ additive reward.}
Under identical settings ($\lambda{=}0.8$, median budget), both multiplicative and additive formulations achieve comparable compression trajectories (Figure~\ref{fig:ablation_curves}(a) and~(b)). The accuracy curves (Figure~\ref{fig:ablation_curves}(c)) also follow similar trends, though the multiplicative variant exhibits slightly less oscillation and achieves modestly higher final performance (82.80\% vs.\ 80.83\% in Table~\ref{tab:ablation}). We attribute this to the tighter objective coupling: the multiplicative form exclusively rewards the desired \textit{correct-and-concise} category, whereas additive combination permits correct-but-verbose trajectories to accumulate positive reward, introducing noise into the optimization landscape.

\paragraph{Effect of $\lambda$.}
We compare $\lambda \in \{0.4, 0.8, 1.0\}$ (all with median budget, multiplicative reward). As shown in Figure~\ref{fig:ablation_curves}(a), all three produce nearly identical compression in the first 10 steps, after which higher $\lambda$ drives faster length reduction. In accuracy (Figure~\ref{fig:ablation_curves}(c)), $\lambda{=}0.8$ peaks earliest (step 20, ${\sim}75\%$) while $\lambda{=}1.0$ and $\lambda{=}0.4$ peak later (step 24 and step 32, at ${\sim}70$--$72\%$) with more oscillation. The slightly lower peak of $\lambda{=}0.4$ reflects its weaker compression pressure, causing the model to converge more slowly; $\lambda{=}1.0$ compresses aggressively but sacrifices some accuracy headroom. We adopt $\lambda{=}0.8$ as a balanced default that achieves the highest accuracy at a moderate compression rate.

\subsection{Analysis}
\label{sec:analysis}

\begin{figure}[h]
    \centering
    \includegraphics[width=\columnwidth]{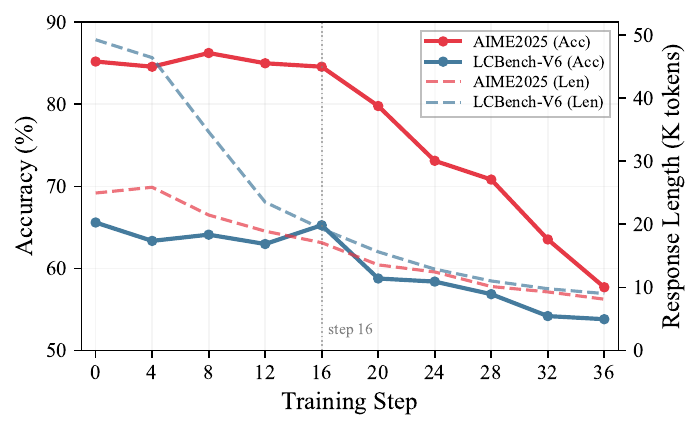}
    \caption{Training dynamics of HMPO (adaptive median) on Qwen3.5-9B. Solid lines (left): accuracy on AIME2025 and LiveCodeBench-V6, stable until step 16. Dashed lines (right): response length drops monotonically (LiveCodeBench reduces 49K$\to$19K by step 16). Dotted line: selected checkpoint.}
    \label{fig:benchmark_curves_9b}
\end{figure}

\paragraph{Training dynamics and checkpoint selection.}
We use AIME 2024 to monitor training progress. As noted in \S\ref{sec:ablation}, early-step validation accuracy is artificially deflated by 24K training truncation. However, we empirically find that the validation compression ratio reliably predicts downstream test compression. To further characterize this trade-off, we evaluate every 4th checkpoint of HMPO on Qwen3.5-9B using full-length inference on two representative benchmarks. As shown in Figure~\ref{fig:benchmark_curves_9b}, several key patterns emerge:
\textbf{(1) Graceful degradation with a clear inflection point:} Accuracy on AIME2025 (red solid line) remains stable through step 16 (85.21\%$\to$84.58\%), then drops sharply beyond step 20 as essential reasoning steps are excessively pruned.
\textbf{(2) Disproportionate compression on verbose tasks:} LiveCodeBench output length (blue dashed line) plummets from 49K to 19K tokens (61\% compression) by step 16 with negligible accuracy loss ($65.60\% \to 65.27\%$), suggesting substantial redundancy in the Base Model's code reasoning.
\textbf{(3) Monotonic length decrease:} Response lengths (dashed lines, right axis) decrease smoothly across both benchmarks, confirming that the adaptive median budget forms a stable compression curriculum without sudden collapses. These dynamics dictate our early-stopping strategy (typically around step 16--20), securing optimal compression before accuracy degrades.
\begin{figure}[t]
    \centering
    \includegraphics[width=\columnwidth]{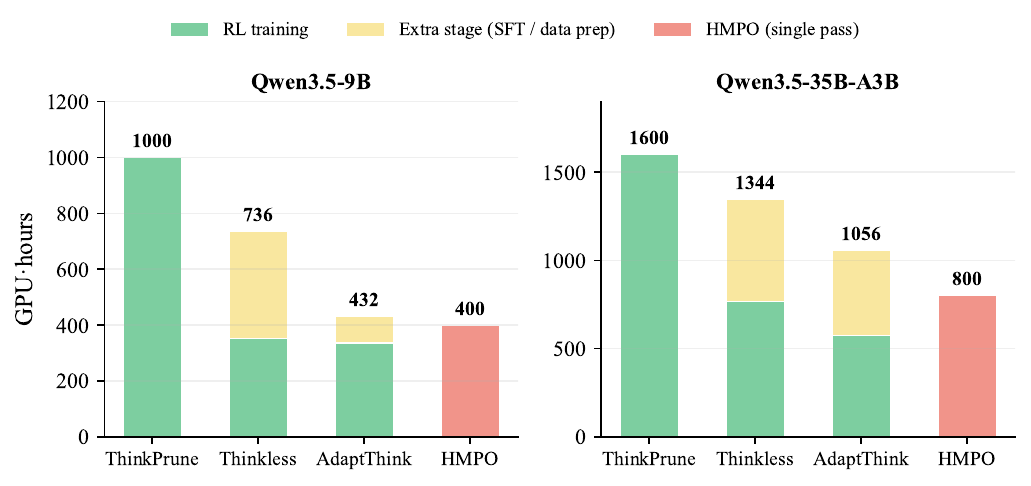}
    \caption{Training cost (GPU$\cdot$hours) comparison. Stacked bars show main training (teal) and extra stages such as SFT or data preprocessing (gold). HMPO (orange) requires only a single-pass GRPO run with no additional stages.}
    \label{fig:gpu_hours}
\end{figure}

\paragraph{Training cost comparison with baselines.}
Figure~\ref{fig:gpu_hours} compares total GPU$\cdot$hours for each method.
ThinkPrune requires 2.5$\times$ HMPO's cost on 9B (1000 vs.\ 400) due to multi-round iterative pruning.
Thinkless incurs 1.8$\times$ overhead from its SFT distillation stage.
AdaptThink has comparable RL cost but adds significant data preprocessing time (96--480 GPU$\cdot$h).
HMPO achieves competitive or superior compression in a single training pass with no auxiliary stages, making it the most cost-effective approach at both model scales.

\section{Conclusion}
In this paper, we introduced HMPO, a cost-effective, single-stage reinforcement learning framework designed to compress verbose Chain-of-Thought (CoT) reasoning. By synergizing a dynamic correct-median budget with a multiplicative cosine-decay reward, HMPO eliminates the need for tedious manual budget tuning and mathematically inoculates the training process against reward hacking. Extensive evaluations demonstrate that HMPO achieves near-optimal accuracy-compression trade-offs. Trained exclusively on mathematical data, the learned concise reasoning patterns generalize robustly across diverse out-of-domain tasks (code, science, instruction following) and scale seamlessly from 9B dense models up to 122B MoE architectures. By completely removing the reliance on computationally prohibitive multi-stage pipelines, SFT warm-ups, and task-specific tuning, HMPO provides a highly practical, drop-in optimization strategy for the efficient real-world deployment of advanced reasoning LLMs.

\section*{Limitations}
\label{sec:limitations}
While HMPO achieves consistent compression with negligible accuracy
loss across multiple benchmarks and model scales, our study has
several limitations that we leave for future work.
\paragraph{Multi-turn reasoning.}
Our evaluation focuses on \emph{single-turn} reasoning settings, where
the model produces one self-contained chain-of-thought per query. We
do not study \emph{multi-turn} scenarios such as interactive problem
solving, follow-up clarification, or dialogue-based tutoring, in which
reasoning unfolds progressively across turns and earlier thoughts may
be reused or revised. ``Concise reasoning'' is more nuanced in such
settings: the model must avoid overthinking within each turn while
also avoiding redundant re-derivations across turns. Extending HMPO
to multi-turn reasoning, e.g.\ by computing the adaptive budget over
a turn-aware aggregation of rollouts, is a promising direction we
leave for future work.
\paragraph{Agentic tasks.}
A related but distinct setting concerns agentic tasks, where the
model interacts with external tools or environments through tool
calls, observations, and intermediate state. We perform only a
preliminary investigation in such settings, and the design of HMPO is
not specifically tailored to them. In agentic settings, reasoning
length must be balanced against tool-call planning, intermediate
state tracking, and recovery from failed actions; a budget that is
appropriate for self-contained reasoning may be too tight for plans
that legitimately require longer deliberation. A systematic study of
HMPO under agentic workflows, including budgets that are conditioned
on tool usage or trajectory phase, remains an important direction.

\bibliography{emnlp}

\section*{Appendix}
\appendix

\section{Training Hyperparameters}
\label{app:hyperparams}

Table~\ref{tab:hyperparams_full} lists the shared training configuration across all models. Table~\ref{tab:parallelism} details per-model parallelism settings.
\begin{table}[h]
\centering
\scriptsize
\caption{Shared HMPO training hyperparameters.}
\label{tab:hyperparams_full}
\begin{tabular*}{\columnwidth}{@{\extracolsep{\fill}}ll@{}}
\toprule
\textbf{Hyperparameter} & \textbf{Value} \\
\midrule
\multicolumn{2}{@{}l}{\textit{RL Algorithm}} \\
Base algorithm & GRPO \\
Rollout group size $G$ & 10 \\
Clip ratio $\epsilon$ (low / high) & 0.2 / 0.2 \\
Clip ratio $c$ & 10.0 \\
KL coeff.\ (reward) & 0.001 \\
KL loss coeff. & 0.001 \\
KL loss type & low-var KL \\
Entropy coeff. & 0 \\
Loss aggregation & token-mean \\
Total epochs & 1 \\
\midrule
\multicolumn{2}{@{}l}{\textit{HMPO Reward}} \\
Budget strategy & adaptive median \\
$\lambda$ & 0.8 \\
Composition & multiplicative \\
\midrule
\multicolumn{2}{@{}l}{\textit{Optimizer}} \\
Optimizer & AdamW \\
Learning rate & $5\!\times\!10^{-7}$ \\
Min LR & $5\!\times\!10^{-8}$ \\
Schedule & cosine decay \\
Warmup & 5\% \\
\midrule
\multicolumn{2}{@{}l}{\textit{Data}} \\
Training data & DeepMath-103K (lv.5--9) \\
Samples & 6.5K \\
Validation & AIME'24 (30 problems.) \\
Max prompt & 4,096 tok. \\
\midrule
\multicolumn{2}{@{}l}{\textit{Checkpointing}} \\
Save / val freq. & every 4 steps \\
Val samples & 4 (do\_sample) \\
\bottomrule
\end{tabular*}
\end{table}

\begin{table}[h]
\centering
\scriptsize
\caption{Per-model parallelism and batch settings.}
\label{tab:parallelism}
\begin{tabular*}{\columnwidth}{@{\extracolsep{\fill}}lccc@{}}
\toprule
& \textbf{9B} & \textbf{35B} & \textbf{122B} \\
\midrule
Nodes$\times$GPUs & 2$\times$8 & 4$\times$8 & 8$\times$8 \\
TP & 2 & 1 & 1 \\
PP & 4 & 8 & 8 \\
CP & 1 & 1 & 1 \\
EP & 1 & 4 & 4 \\
ETP & 1 & 1 & 1 \\
Gen TP & 2 & 8 & 8 \\
Max resp. & 24K & 24K & 24K \\
Batch & 64 & 64 & 128 \\
Mini-batch & 16 & 32 & 64 \\
Mem util. & 0.6 & 0.6 & 0.6 \\
Offload & \checkmark & \checkmark & \checkmark \\
Precision & bf16 & bf16 & bf16 \\
\midrule
\multicolumn{4}{@{}l}{\textit{35B \& 122B MoE settings:}} \\
\multicolumn{4}{@{}l}{\; Attn.\ backend: auto} \\
\multicolumn{4}{@{}l}{\; Recompute: uniform, full, 1 layer} \\
\multicolumn{4}{@{}l}{\; MoE aux loss: 0.01; z-loss: 0.001} \\
\bottomrule
\end{tabular*}
\end{table}
\label{sec:appendix}

\section{Baseline Reproduction Details}
\label{app:baseline_reproduction}

We provide detailed notes on our reproduction of each baseline method. All baselines are trained using Megatron (for SFT stages) and VERL (for RL stages) on their respective original training and validation datasets, strictly following each paper's prescribed training procedure and the hyperparameter configurations specified in their official open-source codebases.

\paragraph{AdaptThink.}
AdaptThink relies on pre-computed per-instance reference statistics from the target model itself. Because reasoning behavior (e.g.,chain-of-thought length and per-instance accuracy) varies substantially across base models of different sizes, these statistics must
be re-computed whenever the base model is changed. Specifically, before training each target model, we sample $K{=}16$ rollout responses per training instance and compute per-instance mean accuracy, which serves as the reward baseline
during RL to normalize advantages across problems of varying difficulty.

Another consequence of switching base models is that their problem-solving capacity within a fixed token budget differs: a smaller model may struggle to produce correct answers within a tight no-think budget, causing the no-think branch to yield near-zero accuracy
and thus preventing the policy from learning meaningful think/no-think decisions. To mitigate this, we adjust the no-think length budget according to model capacity: we use $4$K tokens for Qwen3.5-9B, and increase it to $16$K tokens for Qwen3.5-35B-A3B. This
calibration ensures that the no-think branch remains a viable competing trajectory during RL, so that the policy genuinely learns \emph{when} to skip extended reasoning rather than being implicitly forced to think due to insufficient generation length.

\paragraph{Thinkless.}
Thinkless employs a two-stage pipeline: SFT distillation followed by Decoupled GRPO. We observe that the SFT stage---which trains the model on paired long/short responses with control tokens (\verb|<think>| and \verb|<short>|)---causes substantial degradation of the base model's reasoning capability, a limitation acknowledged in the original paper.

In the subsequent RL stage (trained for over 100 steps), this degradation manifests as severe mode collapse: on Qwen3.5-9B, the model converges almost entirely to short mode (\verb|<short>|) within the first 10 steps, while on Qwen3.5-35B-A3B it collapses to think mode (\verb|<think>|) within the same window. We extensively explored the Decoupled GRPO hyperparameters, including the control token update weight $\alpha \in \{0.001, 0.01, 0.5\}$ and the reward coefficient $\gamma \in \{0.5, 0.8\}$, but were unable to prevent the collapse under any configuration. This underscores a fundamental limitation of the approach: its reliance on high-quality SFT data and sensitivity to RL-stage hyperparameters make it fragile when applied to new base models.

\paragraph{ThinkPrune.}
ThinkPrune employs a multi-round iterative RL procedure that progressively tightens the maximum response length across stages (e.g., 4K$\to$3K$\to$2K in the original paper). A critical design decision is checkpoint selection: the original paper prescribes choosing the checkpoint with the shortest output length that maintains less than 10\% accuracy degradation on the validation set (evaluated every 20 steps).

During reproduction, we encountered a practical challenge: when using the original paper's prescribed \texttt{max\_response\_length} values (4K, 3K, 2K), model outputs are almost always truncated at these limits, causing validation accuracy to collapse to near zero. This renders the checkpoint selection criterion inapplicable, as no checkpoint satisfies the $<$10\% degradation threshold.

To resolve this, we decouple the training truncation from evaluation: we set \texttt{max\_response\_length} to 32K during validation (to allow full-length generation for accurate accuracy measurement), while retaining the target length budget as a truncation threshold only during training rollouts. This engineering modification faithfully reproduces the intended training dynamics while enabling meaningful checkpoint selection.

Even with this fix, we find that ThinkPrune's stop criterion is difficult to satisfy precisely in practice---the method tends to over-compress before the accuracy degradation signal becomes detectable at the 20-step evaluation granularity, resulting in checkpoints that have already crossed the accuracy cliff. This observation indirectly validates the advantage of HMPO's continuous monitoring approach (\S\ref{sec:analysis}), which enables finer-grained early stopping at the optimal accuracy-compression inflection point.

\section{Case Study}
\label{app:aime_case_study}
\begin{figure*}[!t]
\begin{tcolorbox}[
    enhanced,
    title={AIME 2026: Mathematical Reasoning Case (Answer: 178)},
    label={box:case-study-aime},
    colback=gray!5!white,
    colframe=gray!80!black,
    fonttitle=\bfseries\large,
    arc=4pt,
    boxrule=0.8pt,
]
\small

\textbf{Question.}
Find the sum of the 10th terms of all arithmetic sequences of integers
that have first term equal to 4 and include both 24 and 34 as terms.

\medskip

\noindent\textbf{Base Model --- Thinking (35{,}965 chars).}
\begin{quote}
\small
``Define $a_n = 4 + (n-1)d$. Since the terms are integers and
$a_1 = 4$, $d$ must be an integer. From $24 = 4 + (m-1)d$ and
$34 = 4 + (k-1)d$, we get $d \mid 20$ and $d \mid 30$, so
$d \mid \gcd(20, 30) = 10$.

[\ldots]

\textcolor{red}{\textit{Wait, I should pause and check if `arithmetic
sequences of integers' implies something else.}} [\ldots]
\textcolor{red}{\textit{Could $d$ be non-integer rational?}} [\ldots]
\textcolor{red}{\textit{Is it possible that $d$ is negative?}}
[\ldots] If $d < 0$, the terms decrease\ldots\ but $24, 34 > 4$, so
$d$ must be positive. \textcolor{red}{\textit{Wait, what if 24 was
before $a_1$?}} No, $a_1$ is the \emph{first} term. [\ldots]

\textcolor{red}{\textit{Wait, what if the indices are not
$1, 2, \dots$ but can be negative?}} [\ldots]
\textcolor{red}{\textit{Wait, the sequence must contain these terms
in that order, or not necessarily contiguous?}} [\ldots]
\textcolor{red}{\textit{What if the sequence wraps around modulo
something?}} [\ldots]
\textcolor{red}{\textit{Wait, I might have missed a case. What if the
arithmetic sequence is not defined by $a_n = a_1 + (n-1)d$?}} [\ldots]

So $a_{10} = 4 + 9d$, giving $\{13, 22, 49, 94\}$.
Sum $= 13 + 22 + 49 + 94 = 178$.''
\end{quote}

\medskip

\noindent\textbf{Base Model --- Response.}
\begin{quote}
\small
The arithmetic sequences are defined by $a_1 = 4$ and the common
difference $d$. Since the terms are integers, $d$ must be an integer.
[\ldots] For $24, 34$ to be terms, $d$ must be a common divisor of
$20$ and $30$, giving $d \in \{1, 2, 5, 10\}$. The 10th term is
$a_{10} = 4 + 9d \in \{13, 22, 49, 94\}$.
Sum $= 13 + 22 + 49 + 94 = 178$. \quad $\boxed{178}$
\end{quote}

\medskip
\hrule
\medskip

\noindent\textbf{HMPO --- Thinking (8{,}631 chars).}
\begin{quote}
\small
``The $n$-th term is $a_n = a_1 + (n-1)d$ with $a_1 = 4$. Since
$a_2 = 4 + d$ must be an integer, $d \in \mathbb{Z}$.

From $24 = 4 + (m-1)d$ and $34 = 4 + (k-1)d$, subtracting yields
$d \cdot (k-m) = 10$. The terms are increasing, so $d > 0$. [\ldots]
$d$ is a positive divisor of $10$, i.e., $d \in \{1, 2, 5, 10\}$.

[\ldots]

For each $d$, $a_{10} = 4 + 9d \in \{13, 22, 49, 94\}$.
Sum $= 13 + 22 + 49 + 94 = 178$.''
\end{quote}

\medskip

\noindent\textbf{HMPO --- Response.}
\begin{quote}
\small
Let $a_n = a_1 + (n-1)d$ with $a_1 = 4$. Since the terms are integers,
$d \in \mathbb{Z}$. For $24, 34$ to be terms, there exist positive
integers $m, k$ with $(m-1)d = 20$ and $(k-1)d = 30$, so $d$ is a
common divisor of $20$ and $30$. Since $24 > 4$, $d > 0$, and
$\gcd(20, 30) = 10$ gives $d \in \{1, 2, 5, 10\}$. The 10th term is
$a_{10} = 4 + 9d \in \{13, 22, 49, 94\}$, whose sum is
$\boxed{178}$.
\end{quote}

\end{tcolorbox}
\caption{Case study on AIME~2026.}
\label{fig:case-study-aime}
\end{figure*}
As illustrated in Figure~\ref{fig:case-study-aime} and Figure~\ref{fig:case-study-lcb} We present qualitative examples comparing Base Model and HMPO outputs to illustrate how HMPO achieves token compression while preserving correctness. Both examples are drawn from Qwen3.5-9B.

\subsection{AIME 2026: Mathematical Reasoning}


\paragraph{Statistics.}
\begin{itemize}[leftmargin=*,nosep]
    \item \textbf{Base Model}: thinking = 41,349 chars, response = 2,198 chars, total = 43,547 chars.
    \item \textbf{HMPO}: thinking = 7,483 chars, response = 1,560 chars, total = 9,043 chars.
    \item \textbf{Compression}: 79.2\%. Both produce the correct answer $\boxed{178}$.
\end{itemize}

\paragraph{Analysis.}
The Base Model's reasoning trace exhibits several inefficiencies:
\begin{enumerate}[leftmargin=*,nosep]
    \item \textbf{Redundant verification of obvious constraints}: The Base Model spends $\sim$800 characters debating whether $d$ must be an integer, considering fractional $d$ values and ruling them out---a conclusion that follows trivially from ``arithmetic sequences of integers.''
    \item \textbf{Repeated self-questioning}: The thinking trace revisits whether $d$ can be negative multiple times (``Is there any possibility that $d$ is negative and 24, 34 are terms before 4?''), each time reaching the same conclusion that $d > 0$.
    \item \textbf{Over-elaboration of basic steps}: Computing $\gcd(20, 30) = 10$ and listing its divisors is expanded across $\sim$500 characters with intermediate prime factorization steps.
\end{enumerate}

In contrast, HMPO's reasoning is structured as a concise step-by-step outline:
\begin{enumerate}[leftmargin=*,nosep]
    \item Identify constraints $\rightarrow$ formulate $(k_1{-}1)d = 20$, $(k_2{-}1)d = 30$.
    \item Determine $d \mid \gcd(20, 30) = 10$, positive divisors $\{1, 2, 5, 10\}$.
    \item Compute $T_{10} = 4 + 9d$ for each $d$, sum them.
    \item Brief arithmetic verification: $13{+}22{+}49{+}94 = 178$.
\end{enumerate}

HMPO eliminates the redundant self-debate while retaining all logically necessary steps. The final response is also more concise (1,560 vs.\ 2,198 chars) while containing identical mathematical content.

\subsection{LiveCodeBench: Code Generation}

\paragraph{Statistics.}
\begin{itemize}[leftmargin=*,nosep]
    \item \textbf{Base Model}: thinking = 136,169 chars, response = 13,488 chars, total = 149,657 chars.
    \item \textbf{HMPO}: thinking = 44,109 chars, response = 3,595 chars, total = 47,704 chars.
    \item \textbf{Compression}: 68.1\%. Both pass all test cases.
\end{itemize}

\paragraph{Analysis.}
Both models arrive at the same algorithmic solution: $O(N^2)$ DP with convex hull trick (CHT) optimization. However, their reasoning processes differ substantially:

\begin{enumerate}[leftmargin=*,nosep]
    \item \textbf{Problem reformulation}: Both correctly identify that the cost formula uses a prefix sum of \texttt{nums}. The Base Model spends $\sim$5K chars re-reading the problem statement and second-guessing its interpretation before reaching this conclusion; HMPO verifies it against examples in $\sim$1K chars.
    \item \textbf{DP formulation}: The Base Model explores multiple DP formulations ($DP[i]$, $DP[i][m]$, back to $DP[i][m]$ with restructuring) before settling on the correct one. HMPO directly formulates the 2D DP and proceeds to optimization.
    \item \textbf{Convex hull trick derivation}: The Base Model derives CHT from scratch with extensive algebraic manipulation ($\sim$20K chars), while HMPO recognizes the standard pattern (minimizing $mx + b$ with monotone queries) and applies it in $\sim$5K chars.
    \item \textbf{Implementation}: The Base Model's response includes both a lengthy explanation of the approach (10K chars) \emph{and} the code (3.5K chars), with the explanation repeating much of the thinking. HMPO produces a clean code-only response (3.6K chars) without redundant explanation.
\end{enumerate}

The key insight is that HMPO does not sacrifice algorithmic sophistication---it still discovers and implements the optimal $O(N^2)$ CHT solution. It simply avoids the exploration dead-ends, repeated self-correction, and verbose re-explanation that inflate the Base Model's output.





\subsection{Summary of Observations}

Across both domains, HMPO's compression manifests through three consistent mechanisms:
\begin{enumerate}[leftmargin=*,nosep]
    \item \textbf{Elimination of redundant verification}: The Base Model repeatedly re-checks conclusions it has already established. HMPO verifies once and moves forward.
    \item \textbf{Direct problem-solving}: HMPO avoids exploring and discarding suboptimal approaches. It converges on the correct method faster, suggesting the RL training has taught the model to recognize solution patterns more efficiently.
    \item \textbf{Concise final responses}: HMPO produces shorter final answers/code without redundant re-explanation of the reasoning already performed in the thinking block.
\end{enumerate}

These patterns explain why moderate compression (42--46\%) can preserve or even \textit{improve} accuracy: by removing self-contradicting explorations and repetitive verification, the model reduces opportunities for error accumulation while retaining all essential reasoning steps.

\begin{figure*}[!htbp]
\begin{tcolorbox}[
    enhanced,
    title={LiveCodeBench: Code Generation Case},
    label={box:case-study-lcb},
    colback=gray!5!white,
    colframe=gray!80!black,
    fonttitle=\bfseries\large,
    arc=4pt,
    boxrule=0.8pt,
]
\small

\textbf{Problem.}
Divide \texttt{nums} ($n \le 1000$) into subarrays to minimize the
total cost, where the $i$-th subarray $[l, r]$ contributes
$(\texttt{nums}[0]{+}\cdots{+}\texttt{nums}[r] + k \cdot i)
 \times (\texttt{cost}[l]{+}\cdots{+}\texttt{cost}[r])$.

\medskip

\noindent\textbf{Base Model --- Thinking (136{,}169 chars).}
\textit{Direct excerpts; \textcolor{red}{red} marks repeated
re-derivations and self-doubt.}
\begin{quote}
\small
``We use dynamic programming. Let $DP[i][m]$ be the minimum cost to
partition the prefix of length $i$ into $m$ subarrays. The transition
is
$DP[i][m] = \min_{j<i} \big(DP[j][m{-}1]
 + (P_N[i] + k\,m)(P_C[i] - P_C[j])\big)$.''

\medskip
\textcolor{red}{\textit{``Wait, if we define $DP[i]$ as the min cost
for prefix $i$ ending with \emph{some} number of subarrays, that
equation is wrong because the new subarray needs to know $m$.''}}
[\ldots]
\textcolor{red}{\textit{``But wait, does $m$ matter?''}} [\ldots]
\textcolor{red}{\textit{``Wait, $P_C[i]$ and $P_N[i]$ are fixed for a
specific $i$.''}} [\ldots]
\textcolor{red}{\textit{``Wait, $P_N[i]$ and $P_C[i]$ are sequences.''}}
[\ldots]
\textcolor{red}{\textit{``Wait, this is greedy, which might not work.''}}
[\ldots]
\textcolor{red}{\textit{``Wait, look at the constraints again.
$N \le 1000$.''}} [\ldots]

\medskip
\textcolor{red}{\textit{[Re-verifies Example~1 multiple times,
recomputing $(3{+}1{+}1) \times (4{+}6) = 50$ and
$(3{+}1{+}4{+}2) \times 6 = 60$ to confirm the output is $110$.]}}
[\ldots]

\medskip
``The term inside the minimum is of the form $S_j - C_j \cdot X$, where
$S_j = DP[j][m{-}1]$ and $X = P_N[i] + k\,m$. Slopes $-P_C[j]$ are
monotone and queries $X$ are monotone, so we apply the convex hull
trick (CHT) for $O(N^2)$ overall.''
\end{quote}

\medskip

\noindent\textbf{Base Model --- Response.}
\begin{quote}
\small
[\ldots long mathematical exposition ($\sim$10K chars) before code,
duplicating the derivation in the thinking trace, including a full
re-statement of the DP recurrence, the convex hull rewrite, slope
monotonicity, and complexity analysis.]
\medskip

{\footnotesize\ttfamily
class Solution:\\
\hspace*{1em}def minimumCost(self, nums, cost, k):\\
\hspace*{2em}n = len(nums)\\
\hspace*{2em}P\_N, P\_C = [0]*(n+1), [0]*(n+1)\\
\hspace*{2em}for i in range(n):\\
\hspace*{3em}P\_N[i+1] = P\_N[i] + nums[i]\\
\hspace*{3em}P\_C[i+1] = P\_C[i] + cost[i]\\
\hspace*{2em}\#\,\ldots CHT with deque, $\sim$120 lines\,\ldots
}
\end{quote}

\medskip
\hrule
\medskip

\noindent\textbf{HMPO --- Thinking (44{,}109 chars).}
\textit{Direct excerpts; the trace is largely linear.}
\begin{quote}
\small
``Let $P[i]$ and $C[i]$ be the prefix sums of \texttt{nums} and
\texttt{cost}. The $i$-th subarray contributes
$(P[r] + k \cdot i)(C[r] - C[l-1])$.

[\ldots verifies Example~1 once: subarrays $[3,1]$ and $[4]$ give
$50 + 60 = 110$.]

Let $DP[i]$ be the minimum cost for the prefix of length $i$.
Since the cost depends on the subarray index $c$, we add $c$ as an
outer dimension and compute $DP[i][c]$ from $DP[\cdot][c{-}1]$.

The inner minimization is
$\min_{j<i}\big(DP[j][c{-}1] - (P[i] + k\,c)\cdot C[j]\big)$,
which is a line query of the form $-C[j]\,x + DP[j][c{-}1]$ with
$x = P[i] + k\,c$. Since slopes $-C[j]$ and queries $x$ are both
monotone, the convex hull trick gives amortized $O(1)$ per
transition, hence $O(N^2)$ overall.''

\medskip
[\ldots]

\medskip
\textcolor{teal}{\textit{``Final check on time complexity. $O(N^2)$
with $N = 1000$ is $10^6$ operations, well within limits.''}}
\end{quote}

\medskip

\noindent\textbf{HMPO --- Response.}
\begin{quote}
\small
{\footnotesize\ttfamily
class Solution:\\
\hspace*{1em}def minimumCost(self, nums, cost, k):\\
\hspace*{2em}n = len(nums)\\
\hspace*{2em}prefix\_nums = [0]*(n+1)\\
\hspace*{2em}prefix\_cost = [0]*(n+1)\\
\hspace*{2em}for i in range(n):\\
\hspace*{3em}prefix\_nums[i+1] = prefix\_nums[i] + nums[i]\\
\hspace*{3em}prefix\_cost[i+1] = prefix\_cost[i] + cost[i]\\
\hspace*{2em}dp = [float('inf')]*(n+1); dp[0] = 0\\
\hspace*{2em}\#\,\ldots CHT with deque, $\sim$70 lines\,\ldots
}
\end{quote}

\end{tcolorbox}
\caption{Case study on LiveCodeBench (LeetCode Hard)}
\label{fig:case-study-lcb}
\end{figure*}
\end{document}